\pdfoutput=1

\documentclass[11pt]{article}

\usepackage{acl}

\usepackage{times}
\usepackage{latexsym}

\usepackage[T1]{fontenc}

\usepackage[utf8]{inputenc}

\usepackage{microtype}

\usepackage{graphicx}
\usepackage{xcolor}
\usepackage{dsfont}
\usepackage{amsmath,amssymb,amsfonts}
\usepackage{stmaryrd}

\title{Linguistic communication as (inverse) reward design}

\author{
    Theodore R. Sumers \\
    Computer Science \\
    Princeton University \\ \And
  Robert D. Hawkins  \\
  Princeton Neuroscience Institute \\
  Princeton University \\ \And
  Mark K. Ho  \\ 
  Computer Science \\
  Princeton University \\ \AND
  Thomas L. Griffiths \\
  Computer Science, Psychology \\
  Princeton University  \\\And 
  Dylan Hadfield-Menell  \\ 
 EECS, CSAIL \\
  MIT  \\
}

\begin{document}

\maketitle

\begin{abstract}
Natural language is an intuitive and expressive way to communicate reward information to autonomous agents. It encompasses everything from concrete instructions to abstract descriptions of the world. Despite this, natural language is often challenging to learn from: it is difficult for machine learning methods to make appropriate inferences from such a wide range of input.
This paper proposes a generalization of reward design as a unifying principle to ground linguistic communication: speakers choose utterances to maximize expected rewards from the listener's future behaviors. We first extend reward design to incorporate reasoning about unknown future states in a linear bandit setting. We then define a speaker model which chooses utterances according to this objective. Simulations show that short-horizon speakers (reasoning primarily about a single, known state) tend to use instructions, while long-horizon speakers (reasoning primarily about unknown, future states) tend to describe the reward function. We then define a pragmatic listener which performs inverse reward design by jointly inferring the speaker's latent horizon and rewards. Our findings suggest that this extension of reward design to linguistic communication, including the notion of a latent speaker horizon, is a promising direction for achieving more robust alignment outcomes from natural language supervision.

\end{abstract}


\section{Introduction}
Imagine taking up mushroom foraging as a hobby. How would you learn which fungi are delicious and which are deadly? Learning from direct experience~\cite{sutton2018reinforcement} seems risky.
But how might we best learn from others?
Prior work in reinforcement learning (RL) has examined a number of social learning strategies, including passive \emph{inverse reinforcement learning} \citep[observe an expert pick mushrooms, then infer their reward function;][]{ng2000algorithms, abbeel2004apprenticeship} or active \emph{preference learning} \citep[offer an expert pairs of mushrooms, observe which one they eat, and infer their reward function;][]{markant2014better,christiano2017deep, basu2018learning}.

We posit that few humans would rely on such indirect observations if they had access to a cooperative teacher \cite{velez2021learning,gweon2021inferential,wang2020mathematical}. 
For example, an expert guiding a foraging trip might \emph{demonstrate} or verbally \emph{instruct} the learner to pick certain mushrooms rather than others \cite{shafto_pedagogical, ho2016showing}.
While such explicit instruction has been a useful tool for guiding RL agents \cite{goyal2019using,luketina2019survey,fu_2019_goals,Tellex2020}, natural language affords much richer forms of expression. 
For example, an expert teaching a seminar might \emph{describe} how to recognize edible or toxic mushrooms from their features.\footnote{Or write a book on the topic, e.g.~\citet{hyman2021}.}
Descriptive language is particularly powerful if learners can expect experts to prioritize \emph{relevant} and \emph{context-sensitive} information~\cite{sperber1986relevance,tessler2019language}. 


To formalize these expectations, we generalize models of \emph{reward design}~\cite{singh2009rewards} to linguistic communication in a linear bandit setting. Section~\ref{section_reward_design} begins by defining a speaker that chooses utterances to maximize an (imagined) listener's expected rewards over the likely distribution of future states. Section~\ref{section_utterances} shows that speakers focused on a single state prefer instructions (designating an action to take), while those reasoning about many states prefer descriptions (providing information about the reward function). Finally, we consider how a listener might learn from such a speaker. Section~\ref{section_ird} defines a pragmatic listener which performs \emph{inverse} reward design~\cite[IRD,][]{hadfield2017inverse}, to learn about rewards from both instructions and descriptions.

\begin{figure*}[h!]
    \centering
    \includegraphics[width=16cm]{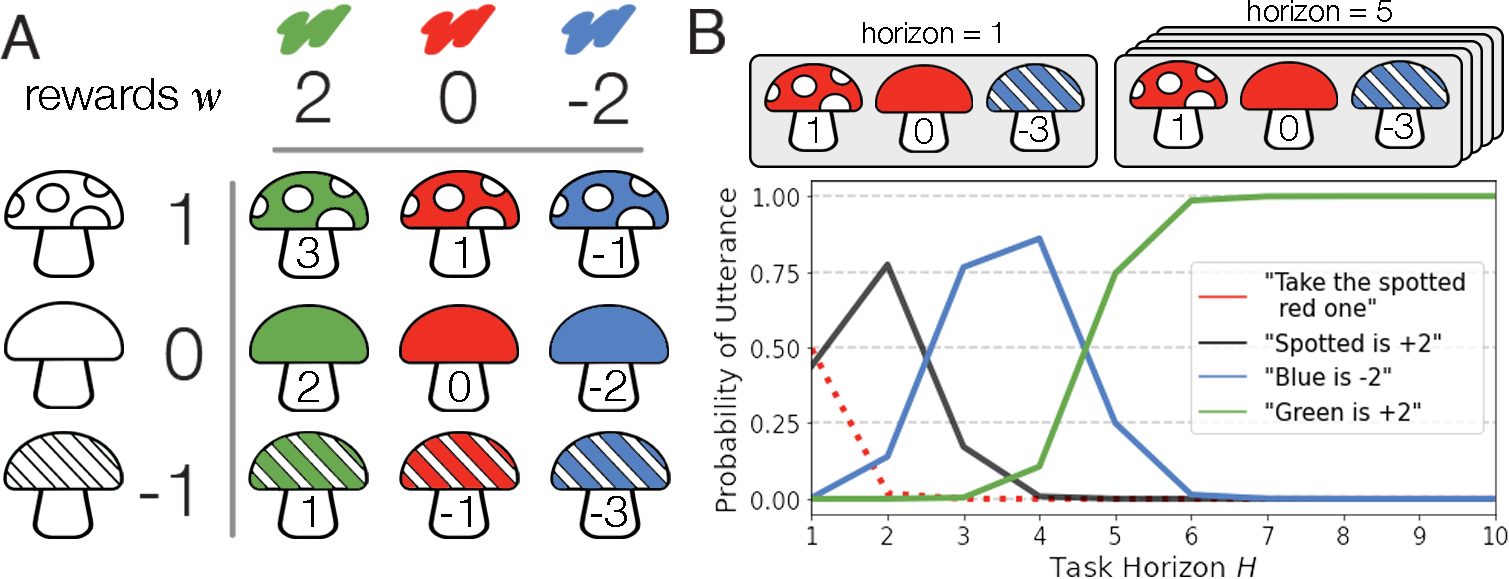} 
    \caption{\textbf{A}: Rewards associated with features determine whether actions (mushrooms) are high or low reward (tasty or toxic). \textbf{B}: Speaker's choice of utterances as a function of horizon $H$ for this start state. At short horizons (maximum supervision), speakers often use instructions or exaggerated descriptions. As the horizon lengthens, there are more unknown states, and speakers prefer truthful descriptions which provide generally useful information. Pragmatic listeners can exploit this pattern to jointly infer a speaker's horizon and rewards.}
    \label{fig:linear_bandits_setup}
\end{figure*}

Using IRD on natural language input offers two distinct benefits over its non-linguistic formulation.
First, language is \emph{expressive} yet tractable (for humans): while reward functions are notoriously difficult to specify~\cite{amodei2016concrete}, natural language provides an accessible and expansive space of proxy rewards.
Second, language can address \emph{future settings}: speakers can refer to actions or features which are not physically present. Thus, while reward design and IRD assume the reward designer optimizes a known Markov Decision Process (MDP), our formulation relaxes this requirement. We show that pragmatic listeners which jointly infer the speaker's reward function and distribution over states reliably outperform a literal listener. 

\section{Communication as Reward Design}
\label{section_reward_design}

\paragraph{Linear Bandits}
We begin by formulating the reward design problem in a \emph{linear bandit} setting~\cite{bandit_algorithms2020, amin2017repeated}. Formally, we define a set of $A$ possible actions. Actions are associated with a binary feature vector $\phi: A \rightarrow \{0, 1\}^K$ (e.g. a mushroom may be green or not; have spots or not). 
Rewards are defined as a function of these features: $R: \phi(a) \rightarrow \mathbb{R}$.
We assume they are a linear combination of the features:
\begin{equation}
    R(a, w) = w^\top\phi(a)
\label{eq_linear_rewards}
\end{equation}
so $w$ is a vector that defines the value of each feature (e.g. green mushrooms are tasty and blue are toxic; see Fig.~\ref{fig:linear_bandits_setup}A). 
Each task consists of a sequence of $H$ i.i.d. states.  
At each time step $t<H$, the agent is presented with a state $s_t$ consisting of a subset of possible actions: $s_t \subseteq A$ (e.g., a particular mushroom patch). 
They choose an action $a \in s_t$ according to their policy, $\pi_L: S \rightarrow \Delta (A)$.

While the bandit problem is typically considered as an individual learning problem, we assume that rewards are not directly observable and instead ask how agents should learn \emph{socially}. 
We formalize the social learning problem by introducing a second agent: a speaker who knows the true rewards $w$ and the initial state $s_0$, and produces an utterance $u$. The listener updates their policy to $\pi_L( a \mid u,s)$ before beginning to choose actions. 
Intuitively, the horizon $H$ determines how much supervision the speaker exerts. $H=1$ is maximum supervision (i.e. guided foraging), whereas $H \to \infty$ is minimal supervision (teaching the listener to forage in future settings). We first assume $H$ is known to both listener and speaker, but later relax this assumption. 

This social learning framework exposes two interrelated problems.
First, what should the speaker agent say to be most helpful? 
And second, how should the listener update their policy in light of this information? 

\paragraph{Speakers as Reward Designers} 
Drawing on the Rational Speech Act framework~\cite[RSA,][]{goodman_2016},  we define a speaker $S_1$ that chooses utterances $u$ according to a utility function $U_{S_1}(\cdot)$:
\begin{equation}
S_1(u) \propto \exp{(\beta_{S_1} \cdot U_{S_1}(u))}
\end{equation}
where $\beta_{S_1}$ is the speaker's soft-max temperature. 

But what utility is appropriate? 
Rather than defining utility simply as Gricean informativeness~\cite{grice1975logic}, i.e. inducing true beliefs, we suggest that a cooperative speaker should \emph{maximize the listener's rewards}, thus grounding utility in terms of the listener's subsequent actions.\footnote{For other recent action-oriented RSA formulations, see~\cite{jiang2021individual, stacy2021modeling, sumers2021extending}.} 


When the state is known, the \emph{present} utility of an utterance is the expected reward from using the resulting policy to choose an action in that state:
\begin{equation}
    U_\text{Present}(u \mid s, w) = \sum_{a \in s} \pi_{L}(a \mid u, s) R(a, w)
\label{eq_local_rewards}
\end{equation}
This formulation is equivalent to the \emph{reward design} objective~\cite{singh2009rewards, hadfield2017inverse}, where the reward designer chooses a proxy reward for a single, known MDP. However, because only the first state is known, we must also consider how well the policy \emph{generalizes} to other mushroom patches. Thus, unlike the reward design objective, speakers may reason about future states. We represent the \emph{future} utility of an utterance with respect to some distribution over states $P(s)$:
\begin{equation}
    U_\text{Future}(u \mid w) = \sum_{s \in S} U_\text{Present}(u \mid s, w)P(s)
\label{eq_expected_rewards}
\end{equation}
Because states are i.i.d. in the bandit setting, a speaker optimizing for a horizon $H$ can be defined as a linear combination of Eqs.~\ref{eq_local_rewards} and~\ref{eq_expected_rewards}:
\begin{equation}
    U_{S_1}(u \mid w, s, H) = U_\text{Present} + (H-1)U_\text{Future}
\label{eq_speaker_combined_utility}
\end{equation}
where $H=1$ reduces to Eq.~\ref{eq_local_rewards}. 
We next define how utterances may affect the listener's policy. 


\section{Choosing Optimal Utterances}
\label{section_utterances}
We formally define two classes of utterances, \emph{instructions} and \emph{descriptions}, by specifying how they affect the policy of a ``literal'' listener. We then show how varying the horizon H systematically affects the speaker's choice of utterance.

\paragraph{Instructions}
Instructions map to specific actions or trajectories~\cite{Tellex2011, jeon2020reward}. Given an instruction, a literal listener executes the corresponding action. If the action is not available, the listener chooses an action randomly:
\begin{equation}
    \pi_{L_0}(a \mid u_{\text{instruct}}, s) =  
\begin{cases}
    \delta_{\llbracket u \rrbracket (a)} & \text{if } a \in s \\
    \frac{1}{\mid s \mid} & \text{if } a \notin s \\
\end{cases}
\label{eq_literal_listener_instruction}
\end{equation}
where $\delta_{\llbracket u \rrbracket(a)}$ represents the meaning of $u$, evaluating to one when utterance $u$ grounds to $a$ and zero otherwise.\footnote{We assume that groundings are known, i.e. the literal listener understands the meaning of utterances.} An instruction is a \emph{partial policy}: it designates the correct action in a subset of states. 

\paragraph{Descriptions}
\label{section_descriptions}
Rather than mapping to a specific action, descriptions provide information about the world~\cite{Ling2017_captions, narasimhan2018grounding, sumers2021learning}. Following~\citet{sumers2021extending}, we assume that descriptions provide the reward of a single feature, similar to feature queries~\cite{basu2018learning}. 

Formally, we define descriptions as a tuple: a one-hot binary feature vector and a scalar value, $\langle\mathds{1}_K, \mathbb{R}\rangle$. These are messages like $\langle$Blue, -2$\rangle$. Given a description, a literal listener ``rules out'' inconsistent hypotheses about reward weights $w$:
\begin{equation}
    L_0(w \mid u_\text{description}) \propto \delta_{\llbracket u \rrbracket (w)} P(w)
\label{eq_literal_listener_belief_update}
\end{equation} 
where $\delta_{\llbracket u \rrbracket(w)}$ represents the meaning of $u$, evaluating to one when $u$ is true of $w$ and zero otherwise. Intuitively, descriptions set $L_0$'s beliefs about the reward of a single feature without affecting others. Descriptions need not be accurate; for example, $\langle$Spotted, +2$\rangle$ is a false but valid utterance.

The listener then marginalizes over possible reward functions to choose an action:
\begin{equation}
    \pi_{L_0}(a \mid u, s) \propto \exp\{\beta_{L_0} \cdot \sum_w R(a, w)L_0(w \mid u))\}
\label{eq_listener_policy}
\end{equation}
where $\beta_{L_0}$ is again a softmax optimality.

\begin{figure}[t!]
    \centering
    \includegraphics[width=7.5cm]{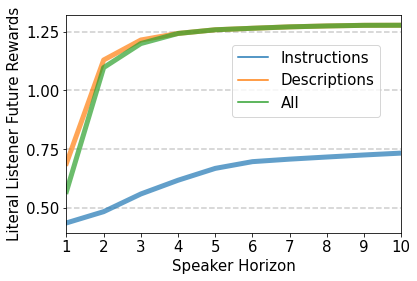}
    \caption{``Future'' rewards (Eq.~\ref{eq_expected_rewards}, averaged over all 84 start states) for a literal listener as a function of horizon and available utterances. At longer horizons, speakers with access to descriptions produce utterances that generalize well.}
    \label{fig:rewards_by_horizon}
\end{figure}

\paragraph{Horizons and Utterance Preferences}
We use simulations to explore the effects of speaker horizons and utterance sets. Fig.~\ref{fig:linear_bandits_setup}A shows our bandit setting. ``Instruction'' utterances correspond to the nine actions. ``Description'' utterances are the $6 \text{ features} \times 5$ values in $[-2, -1, 0, 1, 2]$, yielding 30 feature-value tuples. We assume the listener begins with a uniform prior over reward weights and set $\beta_{L_0} =3, \beta_{S_1} = 10$.\footnote{Because our action space is small (each state has only 3 actions), descriptions are often equivalent to instructions. A lower $\beta_{L_0}$ helps compensate for this.} We use states consisting of three unique actions, giving 84 possible states. 

To quantify how the horizon $H$ affects the generalization of the listener's policy, we repeat the task for all 84 start states using horizons ranging 1-10 and different utterance sets. Fig~\ref{fig:linear_bandits_setup}B shows one example, and Fig~\ref{fig:rewards_by_horizon} plots a literal listener's average future rewards. When the horizon is short (small $H$), speakers focus on the visible state, producing utterances which generalize poorly (low future rewards). As $H$ increases, they provide more generally useful information. Finally, instructions are most useful at short horizons; speakers with access to descriptions use them exclusively when $H>2$. 

\section{Learning from Utterances}
\label{section_ird}
We now ask how the listener should \emph{learn} from the speaker's utterance, using pragmatic inference to recover information about the reward function. 

\begin{figure}[]
    \centering
    \includegraphics[width=7.5cm]{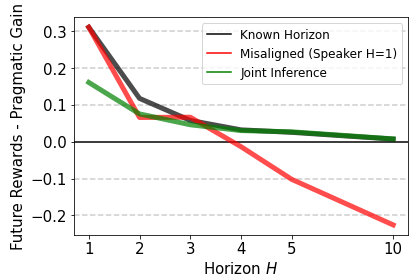}
    \caption{``Future'' reward gain from pragmatic inference (Eq.~\ref{eq_expected_rewards}, $L_1 - L_0$ averaged over all 84 start states). Reward inference works best when the listener knows the speaker's horizon, but can reduce performance if this assumption is incorrect. Jointly inferring the rewards and horizon  (Eq.~\ref{eq_joint_pragmatic_listener}) mitigates this risk.}
    \label{fig:aligment_risk}
\end{figure}

\paragraph{Known Horizon} 
Following the standard RSA formulation, a pragmatic listener $L_1$ can invert the speaker model. When the speaker's horizon $H$ is known, this is equivalent to inverse reward design~\cite{hadfield2017inverse}:
\begin{equation}
    L_1(w \mid s, u, H) \propto S_1(u \mid w, s, H)P(w)
\label{eq_fixed_horizon_pragmatics}
\end{equation}
Given an instruction, $L_1$ can recover information about the reward weights; given a description, $L_1$ can recover information about features that were not mentioned. The $L_1$ listener then chooses actions with respect to this posterior by substituting it into Eq.~\ref{eq_listener_policy}. Fig.~\ref{fig:aligment_risk} shows the gain in ``future'' rewards for a pragmatic listener ($L_1$ - $L_0$) when the speaker has access to both instructions and descriptions, and their horizon is known. Pragmatics are particularly helpful when the speaker has a short horizon and is \emph{not} attempting to provide general information.

\paragraph{Misaligned Horizons} 
However, unlike IRD, in linguistic communication the speaker's horizon $H$ is not explicitly known. Prior work has highlighted the risks of assuming a human is behaving pedagogically when they are not~\cite{milli2020literal}, so we test one form of misalignment: when the speaker $H=1$ but the listener assumes a $H$ ranging from 1-10. Fig.~\ref{fig:aligment_risk} shows that when the pragmatic listener assumes a longer horizon than the speaker intends, it overgeneralizes and performs worse than $L_0$. 

\paragraph{Inference over Speaker Horizons} 
To mitigate the risk of horizon misalignment, we can instead assume the speaker's horizon is unknown. Given an utterance, the listener jointly infers both their horizon and rewards, then marginalizes out the horizon:
\begin{equation}
    L_1(w \mid s, u) \propto \sum_H S_1(u \mid w, s, H)P(H)P(w)
\label{eq_joint_pragmatic_listener}
\end{equation}
We test a pragmatic listener with a uniform prior over $H\in [1, 2, 3, 4, 5, 10]$. This results in more conservative reward inference, but avoids the misalignment risk posed by assuming the speaker's horizon. Fig.~\ref{fig:aligment_risk} shows the results.

\section{Discussion}
\label{section_discussion}
In this work, we formalized communication as reward design, allowing us to unify instructions and descriptions under a single objective. Simulations show that instructions are optimal when the state is known, but descriptions are optimal when considering a distribution over states. Finally, a pragmatic listener can jointly infer the speaker's horizon and reward function.

One important limitation of this work is our reliance on simulations. Future work should validate the speaker model proposed here with behavioral data. Finally, developmental studies indicate that even young children reason about exploration costs when teaching~\cite{bridgers2020young}, suggesting that the reward design objective could be extended further to incorporate reasoning about individual learning. 

\section*{Acknowledgements}
TRS is supported by the NDSEG Fellowship Program and RDH is supported by the NSF (grant \#1911835).
This work was additionally supported by a John Templeton Foundation grant to TLG (\#61454) and a grant from the Hirji Wigglesworth Family Foundation to DHM.

\bibliographystyle{acl_natbib}

\bibliography{references}

\end{document}